\documentclass{article}

\usepackage{microtype}
\usepackage{graphicx}
\usepackage{subcaption}
\usepackage{booktabs} 

\usepackage{hyperref}

\usepackage[preprint]{icml2023}

\usepackage{amsmath}
\usepackage{amssymb}
\usepackage{mathtools}
\usepackage{amsthm}
\usepackage{dsfont}

\usepackage[capitalize,noabbrev]{cleveref}
\usepackage{ulem}
\theoremstyle{plain}
\newtheorem{theorem}{Theorem}[section]

\newtheorem{lemma}[theorem]{Lemma}
\newtheorem{corollary}[theorem]{Corollary}
\newtheorem{fact}[theorem]{Fact}
\theoremstyle{definition}

\theoremstyle{remark}
\newtheorem{remark}[theorem]{Remark}

\usepackage[textsize=tiny]{todonotes}

\newcommand{\R}[0]{\mathbb{R}}

\newcommand{\N}[0]{\mathbb{N}}
\newcommand{\Z}[0]{\mathbb{Z}}

\newcommand{\Prob}[0]{\mathbb{P}}

\newcommand{\Ind}[0]{\mathds{1}}
\newcommand{\eqdef}{\vcentcolon=}

\newcommand\vect[1]{\mathbf{#1}}

\newcommand{\iid}{\stackrel{{\text{i.i.d.}}}{\sim}}

\newcommand\p[1]{\left( {#1}\right)}

\newenvironment{sproof}{%
  \proof}{\endproof}

\icmltitlerunning{Private Statistical Estimation of Many Quantiles}

\begin{document}
\def\UrlBreaks{\do\/\do-}

\twocolumn[
\icmltitle{Private Statistical Estimation of Many Quantiles}
\icmlsetsymbol{equal}{*}

\begin{icmlauthorlist}
\icmlauthor{Clément Lalanne}{1}
\icmlauthor{Aurélien Garivier}{2}
\icmlauthor{Rémi Gribonval}{1}
\end{icmlauthorlist}

\icmlaffiliation{1}{Univ Lyon, EnsL, UCBL, CNRS, Inria,  LIP, F-69342, LYON Cedex 07, France}
\icmlaffiliation{2}{Univ. Lyon, ENS de Lyon, UMPA UMR 5669, 46 allée d'Italie, F-69364 Lyon cedex 07}

\icmlcorrespondingauthor{Clément Lalanne}{clement.lalanne@ens-lyon.fr}

\icmlkeywords{Privacy, Estimation, Quantiles}

\vskip 0.3in
]
\printAffiliationsAndNotice{}
\begin{abstract}
This work studies the estimation of many statistical quantiles under differential privacy. More precisely, given a distribution and access to i.i.d. samples from it, we study the estimation of the inverse of its cumulative distribution function (the quantile function) at specific points. For instance, this task is of key importance in private data generation. We present two different approaches. The first one consists in privately estimating the empirical quantiles of the samples and using this result as an estimator of the quantiles of the distribution. In particular, we study the statistical properties of the recently published algorithm introduced by \cite{kaplan2022differentially} that privately estimates the quantiles recursively. The second approach is to use techniques of density estimation in order to uniformly estimate the quantile function on an interval. In particular, we show that there is a tradeoff between the two methods. When we want to estimate many quantiles, it is better to estimate the density rather than estimating the quantile function at specific points.
\end{abstract}

\section{Introduction}

Computing statistics from real users' data leads to new challenges, notably privacy concerns. Indeed, it is now well documented that the release of statistics computed on them can, without further caution, have disastrous repercussions \cite{narayanan2006break,backstrom2007wherefore,fredrikson2015model,dinur2003revealing,homer2008resolving,loukides2010disclosure,narayanan2008robust,sweeney2000simple,wagner2018technical,sweeney2002k}.
In order to solve this problem, differential privacy (DP) \cite{dwork2006calibrating} has become the gold standard in privacy protection. It adds a layer of randomness in the estimator (i.e. the estimator does not only build on $X_1, \dots, X_n$ but also on another source of randomness) in order to hide each user's data influence. It is notably used by the US Census Bureau~\citep{abowd2018us}, Google~\citep{erlingsson2014rappor}, Apple~\citep{thakurta2017learning} and Microsoft~\citep{ding2017collecting} among others. This notion is properly defined in \Cref{sec:background}, but for now it is only important to view it as a constraint on the estimators that ensures that the observation of the estimator only leaks little information on the individual samples on which it is built on.

Any probability distribution $\Prob$ on $[0, 1]$ is fully characterized by its cumulative distribution function (CDF) defined by 
\begin{equation*}
    F_\Prob(t) \eqdef \Prob \big( (- \infty, t] \big), \quad \forall t \in \R\,.
\end{equation*}
The central topic of this article is the quantile function (QF), $F_\Prob^{-1}$, defined as the generalized inverse of $F_\Prob$:
\begin{equation*}
    F_\Prob^{-1}(p) = \inf \bigg\{ t \in \R \mid p \leq F_\Prob(t) \bigg\}, \quad \forall p \in [0, 1] \;,
\end{equation*}
with the convention $\inf \emptyset = + \infty$. When $\Prob$ is absolutely continuous w.r.t. Lebesgue's measure with a density that is bounded away from $0$, $F_\Prob$ and $F_\Prob^{-1}$ are bijective and are inverse from one another. 

A well-known result is that, under mild hypotheses on $\Prob$, if $U \sim \mathcal{U}([0, 1])$ ($U$ follows a uniform distribution on $[0, 1]$), then $F_\Prob^{-1}(U) \sim \Prob$ \cite{Devr86}. In other words, knowing $F_\Prob^{-1}$ allows to generate data with distribution $\Prob$. It makes the estimation of $F_\Prob^{-1}$ a key component in data generation. Indeed, privately learning the quantile function would then allow generating infinitely many new coherent samples at no extra cost on privacy.

Given $X_1, \dots, X_n \iid \Prob$, this article studies the \emph{private} estimation of $F_\Prob^{-1}(p_j)$ from these samples at prescribed values $\{ p_1, \dots, p_m \} \subset (0, 1)$.
Without privacy and under mild hypotheses on the distribution, it is well-known \cite{van2000asymptotic} that for each $p \in (0, 1)$, the quantity $X_{(E(np))}$ is a good estimator of $F_\Prob^{-1}(p)$,  where $X_{(1)}, \dots, X_{(n)}$ are the order statistic of $X_{1}, \dots, X_{n}$ (i.e. a permutation of the observations such that $X_{(1)}\leq X_{(2)} \leq \dots \leq X_{(n)}$) and $E(x)$ denotes the largest integer smaller or equal to $x$. The quantity $X_{(E(np))}$ is called the empirical (as opposed to statistical) quantile of the dataset $(X_1, \dots, X_n)$ (as opposed to the distribution $\Prob$) of order $p$. 

While the computation of private \emph{empirical} quantiles has led to a rich literature, much less is known on the statistical properties of the resulting algorithms seen as estimators of the \emph{statistical} quantiles of an underlying distribution, compared to more traditional ways of estimating a distribution.

\subsection{Related work}
\label{sec:relatedwork}

Early approaches for solving the private empirical quantile computation used the Laplace mechanism \cite{dwork2006our,dwork2006calibrating} but the high sensitivity of the quantile query made it of poor utility (see \Cref{sec:background} for a quick introduction to differential privacy, including the Laplace mechanism and the notion of sensitity). Smoothed sensitivity-based approaches followed \cite{nissim2007smooth} and managed to achieve greatly improved utility.

The current state of the art for the computation of \emph{a single empirical private quantile} \cite{smith2011privacy} is an instantiation of the so-called exponential mechanism \cite{mcsherry2007mechanism} with a specific utility function (see \Cref{sec:background}) that we will denote QExp (for exponential quantile) in the rest of this article. It is implemented in many DP software libraries \cite{smartnoise,ibmdiffpriv}.

For the computation of \emph{multiple empirical private quantiles}, the problem gets more complicated. Indeed, with differential privacy, every access to the dataset has to be accounted for in the overall privacy budget. Luckily, and part of the reasons why differential privacy became so popular in the first place, composition theorems \cite{dwork2006calibrating,kairouz2015composition,dong2019gaussian,dong2020optimal,abadi2016deep} give general rules for characterizing the privacy budget of an algorithm depending on the privacy budgets of its subroutines. It is hence possible to estimate multiple empirical quantiles privately by separately estimating each empirical quantile privately (using the techniques presented above) and by updating the overall privacy budget with composition theorems. The algorithm IndExp (see \Cref{sec:background}) builds on this framework. However, recent research has shown that such approaches are suboptimal. For instance, \cite{gillenwater2021differentially} presented an algorithm (JointExp) based on the exponential mechanism again, with a utility function tailored for the joint computation of multiple private empirical quantiles directly. JointExp became the state of the art for about a year. It can be seen as a generalization of QExp, and the associated clever sampling algorithm is interesting in itself. Yet, more recently, \cite{kaplan2022differentially} demonstrated that an ingenious use of a composition theorem (as opposed to a more straightforward direct independent application) yields a simple recursive computation using QExp that achieves  the best empirical performance to date. We will refer to their algorithm as RecExp (for recursive exponential). Furthermore, contrary to JointExp, RecExp is endowed with strong utility guarantees \cite{kaplan2022differentially}  in terms of the quality of estimation of the \emph{empirical} quantiles.

In terms of \emph{statistical} utility of the above-mentioned algorithms (i.e. when using the computed private empirical quantiles as statistical estimators of the  statistical quantiles of the underlying distribution), under mild hypotheses, QExp is asymptotically normal \cite{smith2011privacy,asi2020near} and JointExp is consistent \cite{lalanne2022private}.

\subsection{Contributions}

The main contribution of this paper is to obtain concentration properties for RecExp as a private estimator of multiple statistical quantiles (see \Cref{theorem:convergencerecexp}) of a distribution. In order to do so, we adopt a proof framework that controls both the order statistic of $X_1, \dots, X_n$ relatively to the statistical quantiles (see \Cref{lemma:concentrationempiricalquantiles}), and the minimum gap in the order statistic, which is defined as $\min_i X_{(i+1)} - X_{(i)}$, and with the convention $X_{(0)} = 0$ and $X_{(n+1)} = 1$ (see \Cref{lemma:gapsconcentration}). Indeed, this last quantity is of key interest in order to leverage the empirical utility provided by \cite{kaplan2022differentially}. This framework also gives us concentration results for QExp when used to estimate multiple statistical quantiles (see \Cref{theorem:convergenceindexp}). In particular, our results show that when $m$ (the number of statistical quantiles to estimate) is large, RecExp has a much better statistical utility (both in term of proved statistical upper bounds and of experimental behavior) for a given privacy budget than the simple composition of QExp.

We then compare the statistical utility of RecExp to the one of a quantile function built on a simple histogram estimator of the density of $\Prob$. Since this estimator is a functional estimator that estimates all the quantiles in an interval, its statistical utility (see \Cref{theorem:convergencehistogram}) obviously has no dependence on $m$, whereas the utility of RecExp has one. We show that for high values of $m$ the histogram estimator has a better utility than RecExp for a given privacy budget. This theoretical result is confirmed numerically (see \Cref{sec:numericalresults}). For reasonable values of $m$ however, our work consolidates the fact that RecExp is a powerful private estimator, both to estimate \emph{empirical} quantiles of a dataset \cite{kaplan2022differentially} and to estimate the \emph{statistical} quantiles of a distribution (this work). Furthermore, a simple comparison of the upper bounds (\Cref{theorem:convergencerecexp} and \Cref{theorem:convergencehistogram}) can serve as a guideline to decide whether to choose RecExp or an histogram estimator.

\section{Background}
\label{sec:background}

This section presents technical details about differential privacy and private empirical quantiles computation.

\subsection{Differential Privacy}
A randomized algorithm $A$ that takes as input a dataset $(X_1, \dots, X_n)$ (where each $X_i$ lives in some data space, and the size $n$ can be variable) is $\epsilon$-differentially private ($\epsilon$-DP) \cite{dwork2006our,dwork2006calibrating,dwork2014algorithmic}, where $\epsilon > 0$ is a privacy budget, if for any measurable $S$ in the output space of $A$ and any neighboring datasets $(X_1, \dots, X_n) \sim (X_1', \dots, X_{n'}')$ (given some neighboring relation $\sim$) we have
    \[
    \mathbb{P} \big( A(X_1, \dots, X_n) \in S \big) \leq e^\epsilon \times \mathbb{P}\big( A(X_1', \dots, X_{n'}') \in S \big)
    \]
where the randomness is taken w.r.t. $A$. \enlargethispage{0.1cm}

Differential privacy ensures that it is hard to distinguish between two neighboring datasets when observing the output of $A$. The neighboring relation has an impact on the concrete consequences of such a privacy guarantee. A usual goal is to make it hard to tell if a specific user contributed to the dataset. This is typically associated with an "addition/removal" neighboring relation: $(X_1, \dots, X_n) \sim (X_1', \dots, X_{n'}')$ if $(X_1', \dots, X_{n'}')$ can be obtained from $(X_1, \dots, X_n)$ by adding/removing a single element, up to a permutation.
Another choice is the "replacement" neighboring relation: $(X_1, \dots, X_n) \sim (X_1', \dots, X_{n'}')$ if $(X_1', \dots, X_{n'}')$ can be obtained from $(X_1, \dots, X_n)$ up to a permutation by replacing a single entry. 

There are multiple standard ways to design an algorithm that is differentially private. We focus on the ones that will be useful for this article. 

Given a deterministic function $f$ mapping a dataset to a quantity in $\R^d$, the sensitivity of $f$ is
\begin{equation*}
\begin{aligned}
    \Delta f \eqdef \sup_{(X_1, \dots, X_n) \sim (X_1', \dots, X_{n'}')} \big\| &f (X_1, \dots, X_n)  \\ &- f (X_1', \dots, X_{n'}')\big\|_1 \;.
\end{aligned}
\end{equation*}
Given a dataset $(X_1, \dots, X_n)$, the \emph{Laplace mechanism}  returns $f(X_1, \dots, X_n) + \frac{\Delta f}{\epsilon} \text{Lap}(I_d)$ where $\text{Lap}(I_d)$ refers to a random vector of dimension $d$ with independent components that follow a centered Laplace distribution of parameter $1$. This mechanism is $\epsilon$-DP \cite{dwork2014algorithmic}.

If the private mechanism has to output in a general space $O$ equipped with a reference $\sigma$-finite measure $\mu$, one can exploit the \emph{exponential mechanism} \cite{mcsherry2007mechanism} to design it. Given a utility function $u$ that takes as input a dataset $(X_1, \dots, X_n)$ and a candidate output $o \in O$ and returns $u\big( (X_1, \dots, X_n), o \big) \in \R$, which is supposed to measure how well $o$ fits the result of a certain operation that we want to do on $(X_1, \dots, X_n)$ (with the convention that the higher the better), the sensitivity of $u$ is
\begin{equation*}
\begin{aligned}
    \Delta u \eqdef &\sup_{o \in O, (X_1, \dots, X_n) \sim (X_1', \dots, X_{n'}')} \big| u\big((X_1, \dots, X_n), o\big)  \\ &\quad\quad\quad\quad\quad\quad\quad\quad\quad\quad- u\big((X_1', \dots, X_{n'}'), o\big)\big| \;.
\end{aligned}
\end{equation*}
Given a dataset $(X_1, \dots, X_n)$, the exponential mechanism returns a sample $o$ on $O$ of which the distribution of probability has a density w.r.t. $\mu$ that is proportional to $e^{\frac{\epsilon}{2 \Delta u} u\big((X_1, \dots, X_n), o\big) }$. It is $\epsilon$-DP \cite{mcsherry2007mechanism}.

Finally, a simple composition property \cite{dwork2006calibrating} states that if $A_1, \dots, A_k$ are $\epsilon$-DP, $(A_1, \dots, A_k)$ is $k\epsilon$-DP.

\subsection{Private empirical quantile estimation}

This subsection details the algorithms evoked in \Cref{sec:relatedwork} that will be of interest for this article.

\paragraph{QExp.} 

Given $n$ points $X_1, \dots, X_n \in [0, 1]$ and $p \in (0, 1)$, the QExp mechanism, introduced by \cite{smith2011privacy}, is an instantiation of the exponential mechanism w.r.t. $\mu$ the Lebesgue's measure on $[0, 1]$, with utility function $u_{\text{QExp}}$ such that, for any $q \in [0, 1]$,
\begin{equation*}
    u_{\text{QExp}} \big( (X_1, \dots, X_n), q \big) 
    \eqdef 
    - \big| |\{ i | X_i < q\}|- E(np)\big| \;,
\end{equation*}
where for a set, $| \cdot |$ represents its cardinality.
The sensitivity of $u_{\text{QExp}}$ is $1$ for both of the above-mentioned neighboring relations. 
As the density of QExp is constant on all the intervals of the form $(X_{(i)}, X_{(i+1)})$, a sampling algorithm for QExp is to first sample an interval (which can be done by sampling a point in a finite space) and then to uniformly sample a point in this interval. This algorithm has complexity $O(n)$ if the points are sorted and $O(n \log n)$ otherwise. Its utility (as measured by a so-called "empirical error") 
is controlled, cf \cite{kaplan2022differentially} Lemma A.1. This is summarized as follows 
\begin{fact}[Empirical Error of QExp]
\label{fact:empiricalutilityone}
    Consider fixed real numbers $X_1, \dots, X_n \in [0, 1]$ that satisfy $\min_i X_{(i+1)} - X_{(i)} \geq \Delta >0$ with the convention $X_{(0)} = 0$ and $X_{(n+1)} = 1$. Denote $q$ the (random) output of QExp on this dataset, for the estimation of a single empirical quantile of order $p$, and     \begin{equation*}
        \mathfrak{E} \eqdef \Big| \big| \big\{ i | X_i < q \big\} \big| - E(n p) \Big| \;,
    \end{equation*}
    the \emph{empirical error} of QExp.
    For any $\beta \in (0, 1)$, we have
    \begin{equation*}
        \Prob \p{\mathfrak{E} \geq 2 \frac{ \ln \p{\frac{1}{\Delta} } + \ln \p{\frac{1}{\beta}}}{\epsilon}} \leq \beta \;.
    \end{equation*}
\end{fact}
Let us mention that in this article, we use the term \emph{Fact} to refer to results that are directly borrowed from the existing literature in order to clearly identify them. In particular, it is not correlated with the technicality of the result. 

\paragraph{IndExp.} Given $p_1, \dots, p_m \in (0, 1)$, IndExp privately estimates the empirical quantiles of order $p_1, \dots, p_m$ by evaluating each quantile independently using QExp and the simple composition property. Each quantile is estimated with a privacy budget of $\frac{\epsilon}{m}$. The complexity is $O(mn)$ if the points are sorted, $O(mn + n\log n)$ otherwise.

\paragraph{RecExp.} Introduced by \cite{kaplan2022differentially}, RecExp is based on the following idea : Suppose that we already have a private estimate, $q_i$, of the empirical quantile of order $p_i$ for a given $i$. Estimating the empirical quantiles of orders $p_j > p_i$ should be possible by only looking at the data points that are bigger than $q_i$, and similarly for the empirical quantiles of orders $p_j < p_i$.
Representing this process as a tree, the addition or removal of an element in the dataset only affects at most one child of each node and at most one node per level of depth in the tree. The "per-level" composition of mechanisms comes for free in terms of privacy budget, hence only the tree depth matters for composition. By choosing a certain order on the quantiles to estimate, this depth can be bounded by $\log_2 m + 1$. More details can be found in the original article \cite{kaplan2022differentially}.

When using QExp with privacy budget $\frac{\epsilon}{\log_2 m +1}$ for estimating the individual empirical quantiles, RecExp is $\epsilon$-DP with the addition/removal neighborhing relation. This remains valid with the replacement relation if we replace $\epsilon$ by $\epsilon/2$, as the replacement relation can be seen as a two-steps addition/removal relation. RecExp has a complexity of $O(n \log m)$ if the points are sorted and $O(n \log (nm))$ otherwise. The following control of its empirical error is adapted from \cite{kaplan2022differentially} Theorem 3.3.

\begin{fact}[Empirical Error of RecExp]
\label{fact:empiricalutility}
    Consider fixed real numbers $X_1, \dots, X_n \in [0, 1]$ that satisfy $\min_i X_{(i+1)} - X_{(i)} \geq \Delta >0$ with the convention $X_{(0)} = 0$ and $X_{(n+1)} = 1$. Denote $(q_1, \dots, q_m)$ the (random) return of RecExp on this dataset, for the estimation of $m$ empirical quantiles of orders $(p_1, \dots, p_m)$, and
    \begin{equation*}
        \mathfrak{E} \eqdef \max_j \Big| \big|\left\{ i | X_i < q_j \right\}\big| - E(n p_j) \Big| \;,
    \end{equation*}
    the \emph{empirical error} of RecExp. For any $\beta \in (0, 1)$, we have
    \begin{equation*}
    \begin{aligned}
        \Prob &\p{\mathfrak{E} \geq 2(\log_2 m +1)^2 \frac{ \ln \p{\frac{1}{\Delta} } + \ln (m) + \ln \p{\frac{1}{\beta}}}{\epsilon}} \\ 
        &\quad\quad\quad\quad\quad\quad\quad\quad\quad\quad\quad\quad\quad\quad\quad\quad\quad\quad\quad\quad\leq \beta \;.
    \end{aligned}
    \end{equation*}
\end{fact}

\section{Statistical utility of $\star$Exp}
\label{sec:concentration}

\Cref{fact:empiricalutilityone} and \Cref{fact:empiricalutility} control how well QExp, IndExp and RecExp privately estimates \emph{empirical} quantiles of a given dataset. However, they do not tell how well those algorithms behave when the 
dataset is drawn from some probability distribution
and the algorithm output is used to estimate
the \emph{statistical} quantiles of this 
distribution. This is precisely the objective of this section, where we notably highlight the fact that the utility of RecExp scales much better with $m$ (the number of quantiles to estimate) than previous algorithms for this task.

\subsection{How to leverage \Cref{fact:empiricalutilityone} and \Cref{fact:empiricalutility}}

Two difficulties arise when trying to control the \emph{statistical} utility of QExp and IndExp based on  \Cref{fact:empiricalutilityone} and \Cref{fact:empiricalutility}. 

First, the measure of performance (i.e. show mall the empirical error is) controls the deviation w.r.t. the empirical quantiles in terms of \emph{order} :
\begin{equation*}
    \max_j \bigg| \big|\left\{ i | X_i < q_j \right\}\big| - E(n p_j) \bigg| \;.
\end{equation*}
In fact, $E(n p_j) \approx n p_j$ has no link with $F_\Prob$ a priori.
In contrast, from a statistical point of view, the quantity of interest in the deviation w.r.t. the statistical quantiles $(F_\Prob^{-1}(p_1), \dots, F_\Prob^{-1}(p_m))$.
We circumvent that difficulty with the following general purpose lemma :
\begin{lemma}[Concentration of empirical quantiles]
\label{lemma:concentrationempiricalquantiles}
    If $X_1, \dots, X_n \iid \Prob_\pi$ where $\pi$ is a density on $[0, 1]$ w.r.t. Lebesgue's measure such that $\pi \geq \underset{\bar{}}{\pi} \in \R >0$ almost surely, then for any $p \in (0, 1)$ and $\gamma > 0$ such that $\gamma < \min \p{F_X^{-1}(p), 1-F_X^{-1}(p)}$, we have
    \begin{equation*}
\begin{aligned}
    \Prob \Big(\sup_{k \in J} |X_{(E(np) + k)}  &-  F_X^{-1}(p)| > \gamma \Big) \\
    &\leq 2 e^{- \frac{\gamma^2 \underset{\bar{}}{\pi}^2}{8 p} n} + 2 e^{- \frac{\gamma^2 \underset{\bar{}}{\pi}^2}{8 (1-p)} n}\;,
\end{aligned}
\end{equation*}
where 
\begin{equation*}
\begin{aligned}
    J \eqdef \Bigg\{ &\max \p{-E(np) + 1,  - E\p{\frac{1}{2} n \gamma \underset{\bar{}}{\pi}} + 1}, \\ &\dots, \min \p{n-E(np),   E\p{\frac{1}{2} n \gamma \underset{\bar{}}{\pi}} - 1} \Bigg\} \;.
\end{aligned}
\end{equation*}
\end{lemma}
The proof is postponed to \Cref{proofoflemma:concentrationempiricalquantiles}.
The integer set $J$ may be viewed as an error buffer : As long as an algorithm returns a point with an \emph{order} error falling into $J$ (compared to $E(np)$), the error on the \emph{statistical} estimation will be small.

The second difficulty is the need to control the lower bound on the gaps $\Delta$. For many distributions, this quantity can be as small as we want, and the guarantees on the empirical error of QExp, IndExp and RecExp can be made as poor as we want \cite{lalanne2022private}. However, by imposing a simple condition on the density, the following lemma tells that the minimum gap in the order statistic is "not too small".
\begin{lemma}[Concentration of the gaps]
\label{lemma:gapsconcentration}
    Consider $n \geq 1$ and $X_1, \dots, X_n \iid \Prob_\pi$ where $\pi$ is a density on $[0, 1]$ w.r.t. Lebesgue's measure such that $\bar{\pi} \in \R  \geq \pi \geq \underset{\bar{}}{\pi} \in \R >0$ almost surely. 
    Denote $\Delta_i = X_{(i)}-X_{(i-1)}$, $1 \leq i \leq n+1$, with the convention $X_{(0)} = 0$ and $X_{(n+1)} = 1$.
    For any 
    $\gamma >0$ such that $\gamma < \frac{1}{4 \bar{\pi}}$, we have
    \begin{equation*}
        \Prob \p{\min_{i=1}^{n+1} \Delta_i > \frac{\gamma}{n^2}} \geq e^{- 4 \bar{\pi} \gamma} \;.
    \end{equation*}
\end{lemma}
The proof is postponed to \Cref{proofoflemma:gapsconcentration}.

\subsection{Statistical utility of QExp and IndExp}

As a first step towards the analysis of RecExp, and in order to offer a point of comparison, we first build on the previous results to analyze statistical properties of QExp and IndExp.

\begin{theorem}[Statistical utility of QExp]
    \label{theorem:convergenceqexp}
    Consider $n \geq 1$ and $X_1, \dots, X_n \iid \Prob_\pi$ where $\pi$ is a density on $[0, 1]$ w.r.t. Lebesgue's measure such that $\bar{\pi} \in \R  \geq \pi \geq \underset{\bar{}}{\pi} \in \R >0$ almost surely. Denote $q$ the (random) result of QExp on $(X_1, \dots, X_n)$ for the estimation of the quantile of order $p$, where $\min(p,1-p) > 2/n$. For any  $\gamma \in (0,\tfrac{2 \min (p, 1-p)}{\underset{\bar{}}{\pi}})$ 
   \begin{equation*}
        \Prob \big(| q - F_\pi^{-1}(p) | > \gamma\big)
        \leq 
        4 n \sqrt{2 e \bar{\pi}} e^{- \frac{\epsilon n \gamma \underset{\bar{}}{\pi} }{32}}  +  4e^{- \frac{\gamma^2 \underset{\bar{}}{\pi}^2}{8} n} \;.
\end{equation*}
\end{theorem}
\begin{sproof}
    We fix a buffer size $K$ and define $QC$ (for quantile concentration) the event "Any error of at most $K$ points in the order statistic compared to $X_{(E(np))}$ induces an error of at most $\gamma$ on the statistical estimation of $F_\pi^{-1}(p)$". The probability $\Prob \p{QC^c}$ is controlled by \Cref{lemma:concentrationempiricalquantiles}.\\
We fix a gap size $\Delta>0$ and define the event $G$ (for gaps) $\min_i \Delta_i \geq \Delta$, so that
 $\Prob \p{G^c}$ is controlled by \Cref{lemma:gapsconcentration}.\\
    Then, we notice that 
    \begin{equation*}
        \begin{aligned}
            \Prob \big(| q &- F_\pi^{-1}(p) | > \gamma\big)\\
            &\leq \Prob\big(| q - F_\pi^{-1}(p) | > \gamma \big| QC, G\big) 
          + \Prob \p{QC^c} + \Prob \p{G^c} \\
            & \leq
            \Prob \big(\mathfrak{E} \geq K + 1 \big| QC, G\big) 
            + \Prob \p{QC^c} + \Prob \p{G^c} \;,
        \end{aligned}
    \end{equation*}
    where $\mathfrak{E}$ refers to the empirical error of QExp.
    Using \Cref{fact:empiricalutilityone} for a suited $\beta$ controls $\Prob \big(\mathfrak{E} \geq K +1 \bigg| QC, G\big)$. Tuning the values of $K$, $\Delta$ and $\beta$ concludes the proof.
\end{sproof}
The full proof can be found in \Cref{proofoftheorem:convergenceqexp}.

Applying this result to IndExp ($\epsilon$ becomes $\frac{\epsilon}{m}$) together with a union bound gives the following result :
\begin{corollary}[Statistical utility of IndExp]
    \label{theorem:convergenceindexp}
    Consider $n \geq 1$ and $X_1, \dots, X_n \iid \Prob_\pi$ where $\pi$ is a density on $[0, 1]$ w.r.t. Lebesgue's measure such that $\bar{\pi} \in \R  \geq \pi \geq \underset{\bar{}}{\pi} \in \R >0$ almost surely. Denote $\vect{q} \eqdef (q_1, \dots, q_m)$ the (random) result of IndExp on $(X_1, \dots, X_n)$ for the estimation of the quantiles of orders $\vect{p} \eqdef (p_1, \dots, p_m)$, where $\min_i [\min (p_i, 1-p_i)] > 2/n$. For each $\gamma \in \bigg(0,\frac{2 \min_i [\min (p_i, 1-p_i)]}{\underset{\bar{}}{\pi}}\bigg)$ we have
    \begin{equation*}
    \begin{aligned}
        \Prob \big(\| \vect{q} -  F_\pi^{-1}(\vect{p}) \|_\infty > \gamma \big)
       &\leq 
        4 n m\sqrt{2 e \bar{\pi}} e^{- \frac{\epsilon n \gamma \underset{\bar{}}{\pi} }{32 m }}  \\ &+  4 m e^{- \frac{\gamma^2 \underset{\bar{}}{\pi}^2}{8} n} \;,
    \end{aligned}
\end{equation*}
where $F_\pi^{-1}(\vect{p}) = (F_\pi^{-1}(p_1), \dots, F_\pi^{-1}(p_m))$.
\end{corollary}
The proof is postponed to \Cref{proofoftheorem:convergenceindexp}.

So, there exist a polynomial expression $P$ and two positive constants $C_1$ and $C_2$ depending only on the distribution such that, under mild hypotheses,
\begin{equation*}
    \begin{aligned}
        \Prob \big(\| \vect{q} -  &F_\pi^{-1}(\vect{p}) \|_\infty > \gamma \big) \\
       &\leq 
        P(n, m) \max \bigg( e^{- C_1 \frac{\epsilon n \gamma }{m }} , e^{- C_2 \gamma^2 n} \bigg) \;.
    \end{aligned}
\end{equation*}
We factorized the polynomial expression since it plays a minor role compared to the values in the exponential.

\paragraph{Statistical complexity.} The term $P(n, m) e^{- C_2 \gamma^2 n}$ simply comes from the concentration of the empirical quantiles around the statistical ones. It is independent of the private nature of the estimation. It is the price that one usually expects to pay without the privacy constraint.

\paragraph{Privacy overhead.} The term $P(n, m) e^{- C_1 \frac{\epsilon n \gamma }{m }}$ can be called the privacy overhead. It is the price paid for using this specific private algorithm for the estimation. For IndExp, if we want it to be constant, $\epsilon n$ has to roughly scale as $m$ times a polynomial expression in $\log_2 m$. As we will see later in \Cref{theorem:convergencerecexp}, RecExp behaves much better, with $n\epsilon$ having to scale only as a polynomial expression in $\log_2 m$.

A privacy overhead of this type is not only an artifact due to a given algorithm (even if a suboptimal algorithm can make it worse), but in fact a constituent part of the private estimation problem, associated to a necessary price to pay, as captured by several works on generic lower bounds valid for \emph{all} private estimators \cite{duchi2013local,duchi2013localpreprint,acharya2021differentially,acharya2018differentially,AcharyaCST21a,AcharyaCMT21,AcharyaCST21b,AcharyaCFST21,BarnesCO20,BarnesHO20,BarnesHO19,kamath2022improved,butucea2020local,lalanne2022statistical,berrett2019classification,steinberger2023efficiency,kroll2021density}.

\subsection{Statistical properties of RecExp}

With a similar proof technique as in the one of \Cref{theorem:convergenceqexp}, the following result gives the statistical utility of RecExp : 
\begin{theorem}[Statistical utility of RecExp]
    \label{theorem:convergencerecexp}
    Consider $n \geq 1$ and $X_1, \dots, X_n \iid \Prob_\pi$ where $\pi$ is a density on $[0, 1]$ w.r.t. Lebesgue's measure such that $\bar{\pi} \in \R  \geq \pi \geq \underset{\bar{}}{\pi} \in \R >0$ almost surely. Denote $\vect{q} \eqdef (q_1, \dots, q_m)$ the result of RecExp on $(X_1, \dots, X_n)$ for the quantiles of orders $\vect{p} \eqdef (p_1, \dots, p_m)$, where $\min_i [\min (p_i, 1-p_i)]>2/n$.
    For any $\gamma \in (0,\frac{2 \min_i [\min (p_i, 1-p_i)]}{\underset{\bar{}}{\pi}})$ we have
    \begin{equation*}
    \begin{aligned}
        \Prob \big(\| \vect{q} -  F_\pi^{-1}(\vect{p}) \|_\infty > \gamma \big)
        &\leq 
        4 n \sqrt{2e \bar{\pi} m} e^{- \frac{\epsilon  n \gamma \underset{\bar{}}{\pi} }{32 \log_2 (2m)^2}}\\
        &+  4 m e^{- \frac{\gamma^2 \underset{\bar{}}{\pi}^2}{8} n} \;.
    \end{aligned}
\end{equation*}
\end{theorem}
The proof is postponed to \Cref{proofoftheorem:convergencerecexp}.

As with \Cref{theorem:convergenceindexp}, we can simplify this expression as 
\begin{equation*}
    \begin{aligned}
        \Prob \bigg(\| \vect{q} -  &F_\pi^{-1}(\vect{p}) \|_\infty > \gamma \bigg) \\
       &\leq 
        P(n, m) \max \bigg( e^{- C_1 \frac{\epsilon n \gamma }{(\log_2 m)^2 }} , e^{- C_2 \gamma^2 n} \bigg) \;,
    \end{aligned}
\end{equation*}
where $P$ is a polynomial expression and $C_1$ and $C_2$ are constants, all depending only on the distribution.

\paragraph{Statistical complexity.} On the one hand the statistical term of this expression, which is independent of $\epsilon$, is the same as with IndExp. This is natural since the considered statistical estimation problem is unchanged, only the privacy mechanism employed to solve it under a DP constraint was changed.

\paragraph{Privacy overhead.} On the other hand the privacy overhead $P(n, m) e^{- C_1 \frac{\epsilon n \gamma }{(\log_2 m)^2 }}$ is much smaller than the one of IndExp. The scaling of $\epsilon n$ to reach a prescribed probability went from approximately linear in $m$ to roughly a polynomial expression in $\log_2 m$.

In particular and to the best of our knowledge, this scaling in $m$ places RecExp much ahead of its competitors (the algorithms that compute multiple private empirical quantiles) for the task of statistical estimation.

\begin{remark}
    All the results presented in this section require a uniform lower-bound on the density of the distribution from which the data is being sampled. Note that via some minor adaptations in the proofs, all the results can be adapted to the less restrictive hypothesis that the density is lower-bounded on a neighborhood of the statistical quantiles only.
\end{remark}

\section{Uniform estimation of the quantile function}

Private quantile estimators often focus on estimating the quantile function at specific points $p_1, \dots, p_m$ , which is probably motivated by a combination of practical considerations (algorithms to estimate and representing finitely many numbers are easier to design and manipulate than algorithms to estimate a function) and of intuitions about privacy (estimating the whole quantile function could increase privacy risks compared to estimating it on specific points). It is however well-documented in the (non-private) statistical literature that, under regularity assumptions on the quantile function, it can also be approximated accurately from functional estimators, see e.g. \cite{gyorfi2002distribution,tsybakov2003introduction}.


Building on this, this section considers a simple private histogram estimator of the density \cite{wasserman2010statistical} in order to estimate the quantile function in functional infinite norm. This allows of course to estimate the quantile function at $(p_1,\ldots,p_m)$ for arbitrary $m$.
As a natural consequence, we show that when $m$ is very high, for a given privacy level RecExp has suboptimal utility guarantees and is beaten by the guarantees of the histogram estimator. \Cref{theorem:convergencehistogram} and \Cref{theorem:convergencerecexp} give a decision criterion (by comparing the upper bounds) to decide whether to use RecExp or a histogram estimator for the estimation problem.
\subsection{Motivation: lower bounds for IndExp and RecExp}
\label{sec:lowerbounds}

Lower-bounding the density of the exponential mechanism for $u_{\text{QExp}}$ gives a general lower-bound on its utility:
\begin{lemma}[Utility of QExp; Lower Bound]
    \label{lemma:lowerboundqexp}
    Let $X_1, \dots, X_n \in [0, 1]$. Denoting by $q$ the result of QExp on $(X_1, \dots, X_n)$ for the quantile of order $p$, we have for any $t \in [0, 1]$ and any positive $\gamma \in (0,\tfrac{1}{4}]$,
   \begin{equation*}
    \begin{aligned}
        \Prob \bigg(| q - t | > \gamma\bigg)
        &\geq 
        \frac{1}{2} e^{- \frac{n \epsilon}{2}}
         \;.
    \end{aligned}
\end{equation*}
\end{lemma}
Note that this holds without any constraint relating $p$,$n$, or $\gamma$.
The proof is postponed to \Cref{proofoflemma:lowerboundqexp}.
As a consequence, if the points $X_1, \dots, X_n$ are randomized, the probability that QExp makes an error bigger than $\gamma$ on the estimation of a quantile of the distribution is at least $\frac{1}{2} e^{- \frac{n \epsilon}{2}}$.
A direct consequence is that for any $\gamma \in (0,\tfrac{1}{4}]$, the statistical utility of IndExp has a is lower-bounded:
\begin{equation*}
    \Prob \bigg(\| \vect{q} -  F_\pi^{-1}(\vect{p}) \|_\infty > \gamma \bigg)
    \geq \frac{1}{2} e^{- \frac{n \epsilon}{2m}} \;,
\end{equation*}
and the statistical utility of RecExp is also lower-bounded:
\begin{equation*}
    \Prob \bigg(\| \vect{q} -  F_\pi^{-1}(\vect{p}) \|_\infty > \gamma \bigg)
    \geq \frac{1}{2} e^{- \frac{n \epsilon}{2 (\log_2 m +1)}} \;.
\end{equation*}
These are consequences of lower-bounds on the estimation error of the first statistical quantile estimated by each algorithm in its respective computation graph (with privacy level $\epsilon/m$ for IndExp; $\epsilon/(\log_2 m+1)$ for RecExp).

In particular, for both algorithms, utility becomes arbitrarily bad when $m$ increases. This is not a behavior that would be expected from any optimal algorithm. The rest of this section studies a better estimator for high values of $m$.

\subsection{Histogram density estimator}

The histogram density estimator is a well-known estimator of the density of a distribution of probability. Despite its simplicity, a correct choice of the bin size can even make it minimax optimal for the class of Lipschitz densities. 

Under differential privacy, this estimator was first adapted and studied by \cite{wasserman2010statistical}. It is studied both in terms of integrated squared error and in Kolmogorov-Smirnov distance. In the sequel, we need a control in infinite norm. We hence determine the histogram concentration properties for this metric.

Given a a bin size $h > 0$ that satisfies $\frac{1}{h} \in \N$, we partition $[0, 1]$ in $\frac{1}{h}$ intervals of length $h$. The intervals of this partition are called the bins of the histogram. Given $\frac{1}{h}$ i.i.d. centered Laplace distributions of parameter $1$,  $\p{\mathcal{L}_b}_{b \in \text{bins}}$, we define $\hat{\pi}^{\text{hist}}$, an estimator of the supposed density $\pi$ of the distribution as: for each $t \in [0, 1]$,
\begin{equation*}
    \hat{\pi}^{\text{hist}} (t) \eqdef \sum_{b \in \text{bins}} \Ind_b(t) \frac{1}{nh} \p{\sum_{i=1}^n \Ind_b(X_i) + \frac{2}{\epsilon} \mathcal{L}_b} \;.
\end{equation*}
The function that, given the bins of a histogram, counts the number of points that fall in each bin of the histogram has a sensitivity of $2$ for the replacement neighboring relation. Indeed, replacing a point by another changes the counts of at most two (consecutive) bins by one. Hence, the construction of the Laplace mechanism ensures that $\hat{\pi}^{\text{hist}}$ is $\epsilon$-DP.

Note that, as a common practice, we divided by $n$ freely in terms of privacy budget in the construction of $\hat{\pi}^{\text{hist}}$. This is possible because we work with the replacement neighboring relation. The size $n$ of the datasets is fixed and is a constant of the problem.

The deviation between $\pi$ and $\hat{\pi}^{\text{hist}}$ can be controlled.
\begin{lemma}[Utility of $\hat{\pi}^{\text{hist}}$; Density estimation]
\label{lemma:histogramconvergencedensity}
Consider $X_1, \dots, X_n \iid \Prob_\pi$ where $\pi$ is a density on $[0, 1]$ w.r.t. Lebesgue's measure such that $\pi$ is $L$-Lipschitz for some positive constant $L$, and the private histogram density estimator $\hat{\pi}^{\text{hist}}$ with bin size $h$. For any $\gamma > Lh$, we have
    \begin{equation*}
    \Prob \p{\|\hat{\pi}^{\text{hist}} - \pi \|_{\infty} > \gamma} \leq \frac{1}{h} e^{- \frac{\gamma h n \epsilon}{4}} + \frac{2}{h} e^{- \frac{h^2 (\gamma - Lh)^2}{4} n}  \;.
\end{equation*}
\end{lemma}
The proof is postponed to \Cref{proofoflemma:histogramconvergencedensity}.

\subsection{Application to quantile function estimation}

\begin{figure*}[h!]
    \begin{subfigure}{.24\textwidth}
        \centering
        \includegraphics[width=\linewidth]{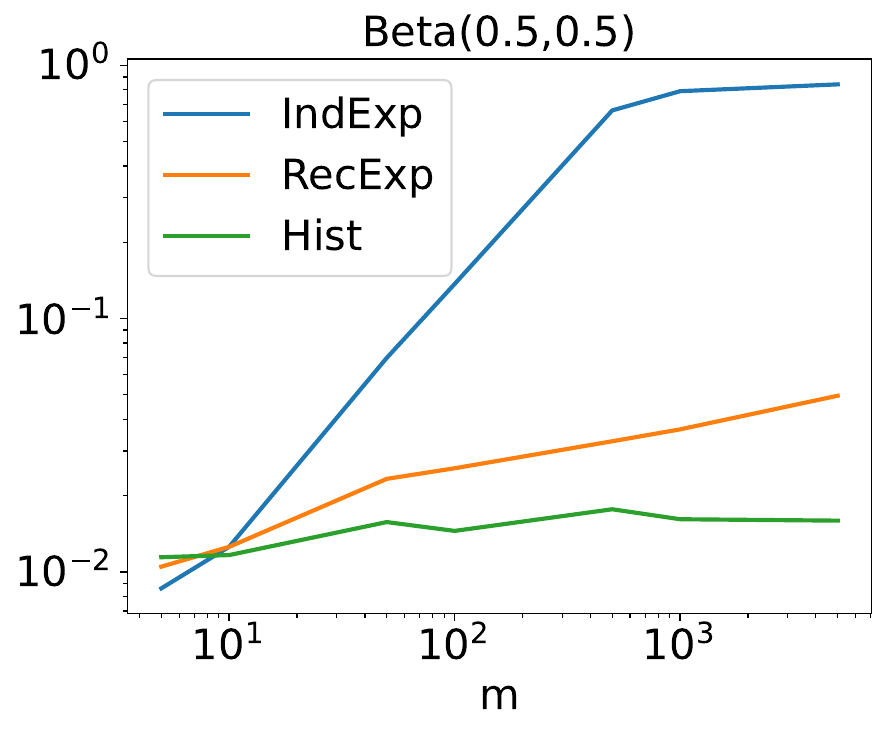}  
        \label{fig:sub-first}
    \end{subfigure}
    \begin{subfigure}{.24\textwidth}
        \centering
        \includegraphics[width=\linewidth]{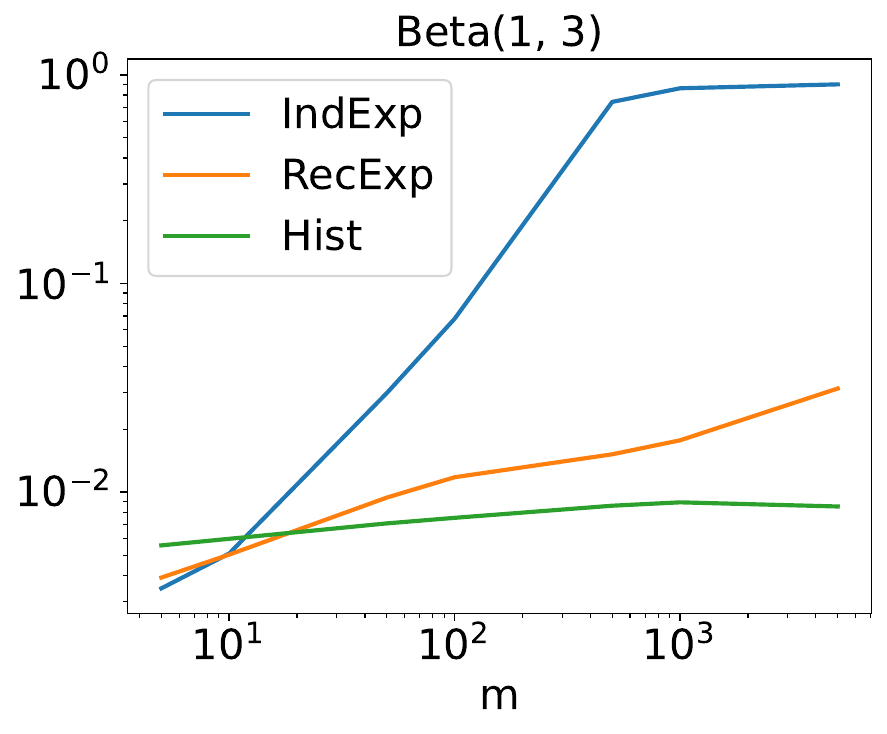}  
        \label{fig:sub-second}
    \end{subfigure} 
    \begin{subfigure}{.24\textwidth}
        \centering
        \includegraphics[width=\linewidth]{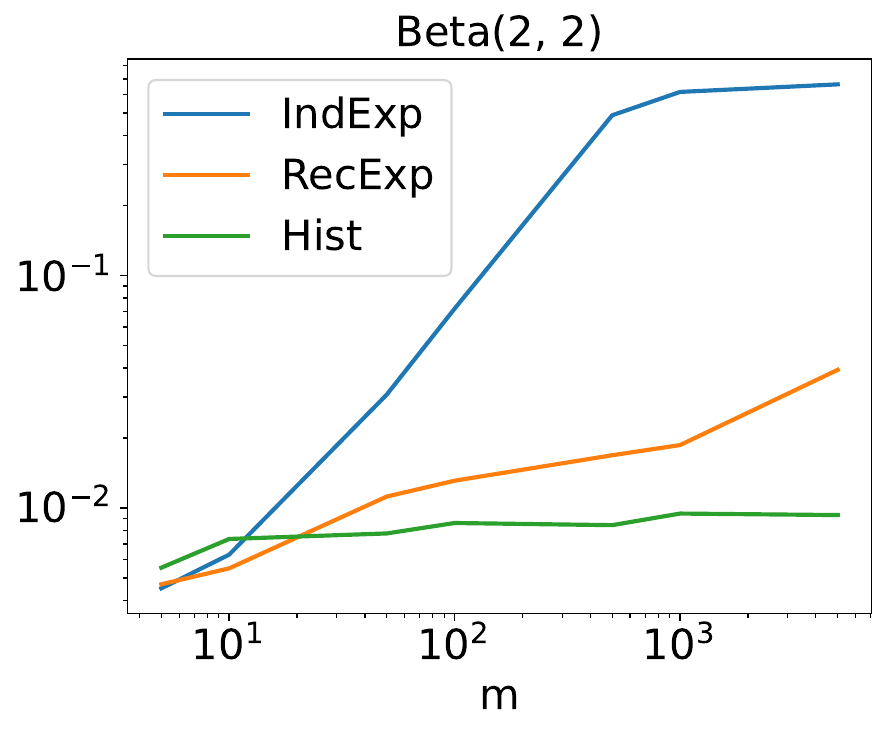}  
        \label{fig:sub-second}
        \end{subfigure}
    \begin{subfigure}{.24\textwidth}
        \centering
        \includegraphics[width=\linewidth]{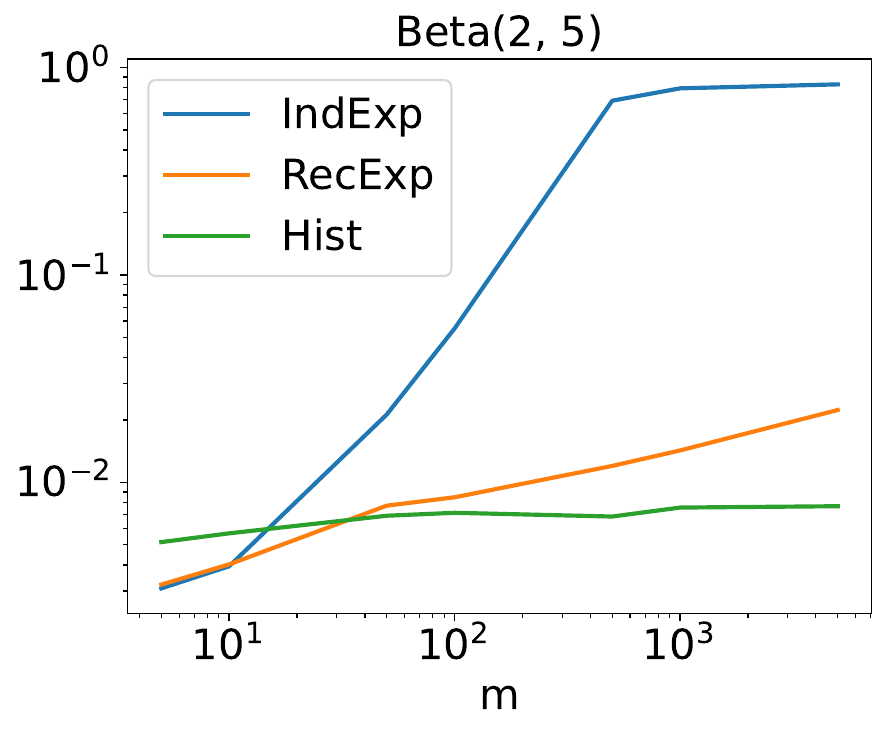}  
        \label{fig:sub-second}
    \end{subfigure}\\
    \centering
    The vertical axis reads the error $\mathbb{E} \left( \|\hat{\mathbf{q}} - F^{-1}(\mathbf{p})\|_\infty \right)$ where $\mathbf{p} = \left( \frac{1}{4}+ \frac{1}{2(m+1)}, \dots, \frac{1}{4}+\frac{m}{2(m+1)}\right)$ for different values of $m$, $n = 10000$, $\epsilon=0.1$, $\hat{\mathbf{q}}$ is the private estimator, and $\mathbb{E}$ is estimated by Monte-Carlo averaging over $50$ runs. The histogram is computed on $200$ bins.
    \caption{Numerical performance of the different private estimators}
    \label{fig:utility}
    \end{figure*}

In order to use $\hat{\pi}^{\text{hist}}$ as an estimator of the quantile function, we need to properly define a quantile function estimator associated with it. Indeed, even if $\hat{\pi}^{\text{hist}}$ estimates a density of probability, it does not necessary integrate to $1$ and can even be negative at some locations. 
Given any integrable 
function $\hat{\pi}$ on $[0,1$], we define its generalized quantile function 
\begin{equation*}
     \quad F^{-1}_{\hat{\pi}}(p) = \inf \left\{ q \in [0, 1] | \int_0^q \hat{\pi} \geq p\right\}, \forall p \in [0, 1] \;,
\end{equation*}
 with the convention $\inf \emptyset = 1$.
 Even if this quantity has no reason to behave as a quantile function, the following lemma tells that $F^{-1}_{\hat{\pi}}$ is close to an existing quantile function when $\hat{\pi}$ is close to its corresponding density.
\begin{lemma}[Inversion of density estimators]
\label{lemma:densityesttoqfest}
Consider a density $\pi$ on $[0, 1]$ w.r.t. Lebesgue's measure such that $\pi \geq \underset{\bar{}}{\pi} \in \R >0$ almost surely.
    If $\hat{\pi}$ is an integrable 
    function that satisfies $\left\| \hat{\pi} - \pi \right\|_{\infty}  \leq \alpha$, and if $p \in [0,1]$ is such that $\left[ F^{-1}_\pi (p) - \frac{2 \alpha}{\underset{\bar{}}{\pi}}, F^{-1}_\pi (p) + \frac{\alpha}{\underset{\bar{}}{\pi}} \right] \subset (0, 1)$, then
    \begin{equation*}
        \left| F^{-1}_\pi (p) - F^{-1}_{\hat{\pi}} (p) \right| \leq \frac{2 \alpha}{\underset{\bar{}}{\pi}} \;.
    \end{equation*}
\end{lemma}
The proof is in \Cref{proofoflemma:densityesttoqfest}.

A direct consequence of \Cref{lemma:histogramconvergencedensity} and \Cref{lemma:densityesttoqfest} is \Cref{theorem:convergencehistogram}. It controls the deviation of the generalized quantile function based on $\hat{\pi}^{\text{hist}}$ to the true quantile function.
\begin{theorem}[Utility of $F_{\hat{\pi}^{\text{hist}}}^{-1}$; Quantile function estimation]
\label{theorem:convergencehistogram}
    Consider $X_1, \dots, X_n \iid \Prob_\pi$ where $\pi$ is a density on $[0, 1]$ w.r.t. Lebesgue's measure such that $\pi$ is $L$-Lipschitz for some positive constant $L$ and that $\pi \geq \underset{\bar{}}{\pi} \in \R >0$ almost surely, 
    and $h< \underset{\bar{}}{\pi}/(4L)$ such that $\tfrac{1}{h} \in \N$. Let $F_{\hat{\pi}^{\text{hist}}}^{-1}$ be
    the quantile function estimator associated with the private histogram density estimator $\hat{\pi}^{\text{hist}}$ with bin size $h$.
    Consider $\gamma_0 \in (2Lh/\underset{\bar{}}{\pi},1/2)$,
    $I \eqdef F_\pi \big( (\gamma_0, 1-\gamma_0) \big)$, and $\|\cdot \|_{\infty, I}$ the sup-norm of functions on the interval $I$. We have
    \begin{equation*}
        \begin{aligned}
            \Prob \bigg( & \|F_{\hat{\pi}^{\text{hist}}}^{-1} - F_{\pi}^{-1}\|_{\infty, I} > \gamma \bigg) \\ &\leq  \frac{1}{h} e^{- \frac{\gamma \underset{\bar{}}{\pi} h n \epsilon}{8}} + \frac{2}{h} e^{- \frac{h^2 }{4} \p{\frac{\gamma \underset{\bar{}}{\pi}}{2} - Lh}^2 n};,
        \end{aligned} 
    \end{equation*}
    whenever $\gamma \in (2Lh/\underset{\bar{}}{\pi},\gamma_0)$.
\end{theorem}
The proof is postponed to \Cref{proofoftheorem:convergencehistogram}.

\paragraph{Analysis of \Cref{theorem:convergencehistogram}.}

As with \Cref{theorem:convergenceqexp} and \Cref{theorem:convergencerecexp}, the upper-bound provided by \Cref{theorem:convergencehistogram} can be split in two terms : The error that one usually expects without privacy constraint, $\frac{2}{h} \exp(- \frac{h^2 }{4} (\frac{\gamma \underset{\bar{}}{\pi}}{2} - Lh)^2 n)$, and the one that come from the private algorithm, $ \frac{1}{h} \exp(- \frac{\gamma \underset{\bar{}}{\pi} h n \epsilon}{8})$. The assumption $\frac{\gamma \underset{\bar{}}{\pi}}{2} > Lh$ ensures that the bin size $h$ and the desired level of precision $\gamma$ are compatible.

\paragraph{Computational aspects.} $\hat{\pi}^{\text{hist}}$ is constant on each bin. Hence, it can be stored in a single array of size $\frac{1}{h}$. If the data points are sorted, this array can be filled with a single pass over all data points and over the array. Then, given $p_1, \dots, p_m \in (0, 1)$ sorted, estimating $F_{\hat{\pi}^{\text{hist}}}^{-1}(p_1), \dots, F_{\hat{\pi}^{\text{hist}}}^{-1}(p_m)$ can be done with a single pass over $p_1, \dots, p_m$ and over the array that stores $\hat{\pi}^{\text{hist}}$. Indeed, it is done by "integration" of the array until the thresholds of the $p_i$'s are reached. The overall complexity of this procedure is $O\p{n + m + \frac{1}{h}}$ to which must be added $O(n \log n)$ if the data is not sorted and $O(m \log m)$ if the targeted quantiles $p_i$ are not sorted.

\paragraph{Comparison with RecExp.}

Comparing this histogram-based algorithm to RecExp is more difficult than comparing RecExp to IndExp. First of all, the results are qualitatively different. Indeed, RecExp estimates the quantile function on a finite number of points and the histogram estimator estimates it on an interval. The second result is stronger in the sense that when the estimation is done on an interval, it is done for any finite number of points in that interval. However, the error of RecExp for that finite number of points may be smaller than the one given by the histogram on the interval. Then, the histogram depends on a meta parameter $h$. With a priori information on the distribution, it can be tuned using  \Cref{theorem:convergencehistogram}. Aditionally, the hypothesis required are different : \Cref{theorem:convergencerecexp} does not require the density to be Lipschitz contrary to \Cref{theorem:convergencehistogram}. Finaly, we can observe that the histogram estimator is not affected by the lower bounds described in \Cref{sec:lowerbounds}. Hence, when all the hypotheses are met, there will obviously always be a number $m$ of targeted quantiles above which it is better to use histograms. The two algorithms are numerically compared in \Cref{sec:numericalresults}.

\begin{remark}
    Notice that the hypothesis of Lipschitzness of the density is only useful for the histogram estimators. In particular the guarantees of RecExp of \Cref{sec:concentration} do not require such hypothesis. This section thus presented a \emph{strict subclass} of the problem on which RecExp may be suboptimal.
\end{remark}

\begin{remark}
    We would like to highlight the fact that histograms are used as an illustration of the suboptimality of RecExp on some instances of the problem. In particular, it does not imply that they are the state of the art on  such instances. It is very possible that other mechanisms perform well in such cases \cite{DBLP:conf/focs/BlockiBDS12,DBLP:journals/corr/abs-2201-03380}. In fact, provided that the inversion from the cumulative distribution function of the distribution to its quantile function is easy (which is typically the case when the density is uniformly lower-bounded), we expect that many private CDF estimators will behave similarly or better on these specific instances \cite{DBLP:conf/focs/BunNSV15,DBLP:conf/colt/KaplanLMNS20,10.1093/jssam/smac021,DBLP:conf/nips/DenisovMRST22,DBLP:journals/corr/abs-2202-11205}.
\end{remark}

\section{Numerical results}
\label{sec:numericalresults}

For the experiments, we benchmarked the different estimators on beta distributions, as they allow to easily tune the Lipschitz constants of the densities, which is important for characterizing the utility of the histogram estimator. 

\Cref{fig:utility} represents the performance of the estimator as a function of $m$. We estimate the quantiles of orders $\mathbf{p} = \left( \frac{1}{4}+ \frac{1}{2(m+1)}, \dots, \frac{1}{4}+\frac{m}{2(m+1)}\right)$ since it allows us to stay in the regions where the density is not too small.

\paragraph{IndExp \emph{vs} RecExp \emph{vs} Histograms.}

\Cref{fig:utility}, confirms our claims about the scaling in $m$ of IndExp and RecExp. Indeed, even if IndExp quickly becomes unusable, RecExp stays at a low error until really high values of $m$. The conclusions of \Cref{sec:lowerbounds} also seem to be verified : Even if RecExp performs well for small to intermediate values of $m$, there is always a certain value of $m$ for which it becomes worse than the histogram estimator. This shift of regime occurs between $m \approx 10$ for the distribution Beta(0.5, 0.5) and $m \approx 40$ for the distribution Beta(2, 5).

\paragraph{Error of the histogram-based approach.} The shape of the error for the histogram estimator is almost flat. Again, it is compatible with \Cref{theorem:convergencehistogram} : The control in infinite norm is well suited for the histograms.

\paragraph{Role of the Lipschitz constant.}

By crossing the shape of the beta distributions (see \Cref{sec:distributions}) and \Cref{fig:utility}, a pattern becomes clear :  The distributions on which the histogram estimator performs best (i.e. the distributions on which it becomes the best estimator for the lowest possible value of $m$) are the distributions with the smallest Lipschitz constant. This was expected since the guarantees of utility of  \Cref{theorem:convergencehistogram} get poorer the higher this quantity is.

\section{Conclusion}

Privately estimating the (statistical) quantile function of a distribution has some interesting properties. For low to mid values of $m$, this article demonstrated that there is a real incentive in estimating it on a finite sample of $m$ points. This was done by using algorithms recently introduced in order to estimate the \emph{empirical} quantiles of a dataset. However, when the number $m$ becomes too high, the previously-mentioned algorithms become suboptimal. It is then more effective to estimate the density with a histogram. Furthermore, the utility results are qualitatively stronger : The estimation is uniform over an interval, as opposed to pointwise on a finite set. \Cref{theorem:convergencerecexp} and \Cref{theorem:convergencehistogram} can be used to decide what method to choose.

An interesting question would be to know if it is possible to modify RecExp in such regimes in order to bridge the gap with histograms. Possibly by adapting the privacy budget to the depth in the computation tree.

Another interesting question would be to investigate the possible (minimax) optimality of the techniques of this article on restricted classes of distributions or regimes of $m$.

\section*{Acknowledgments}
Aurélien Garivier acknowledges the support of the Project IDEXLYON of the University of Lyon, in the framework of the Programme Investissements d'Avenir (ANR-16-IDEX-0005), and Chaire SeqALO (ANR-20-CHIA-0020-01). 

This project was supported in part by the AllegroAssai ANR project ANR-19-CHIA-0009.

Clément Lalanne would like to thank Adam D. Smith for the suggestion of important references.

Additionally, the authors would like top thank the anonymous reviewers for their suggestions and inputs that helped to improve the current version of this article.

\bibliography{biblio}
\bibliographystyle{icml2023}

\newpage
\appendix
\onecolumn

\section{Proof of \Cref{lemma:concentrationempiricalquantiles}}
\label{proofoflemma:concentrationempiricalquantiles}

We define
\begin{equation*}
    \bar{N} \eqdef \sum_{i=1}^n \Ind_{(F_X^{-1}(p) + \gamma, +\infty)} (X_i) \;.
\end{equation*}
Let $k \in \{ -E(np)+1, \dots, n - E(np)\}$. We have the following event inclusion:
\begin{equation*}
    \p{X_{(E(np) + k)} > F_X^{-1}(p) + \gamma}
    \subset \p{\bar{N} \geq  n - (E(np) + k)} 
    \subset \p{\bar{N} \geq  n (1-p) - k - 1} \;. 
\end{equation*}
$\bar{N}$ being a sum of independent Bernoulli random variables, we introduce $\eta \eqdef 1 - p - \gamma  \underset{\bar{}}{\pi}$, a natural upper bound on the probability of success of each of these Bernoulli random variables. Hence, by multiplicative Chernoff bounds, whenever $\frac{\gamma \underset{\bar{}}{\pi}}{\eta} - \frac{k+1}{n \eta} \geq 0$, which is equivalent to $k \leq n \gamma \underset{\bar{}}{\pi} - 1$,
\begin{equation*}
\begin{aligned}
    \Prob \p{X_{(E(np) + k)} > F_X^{-1}(p) + \gamma} 
    &\leq \Prob \p{\bar{N}  \geq n \eta \p{1+ \frac{\gamma \underset{\bar{}}{\pi}}{\eta} - \frac{k+1}{n \eta}}} \\
    &\leq e^{- n \eta {\p{\frac{\gamma \underset{\bar{}}{\pi}}{\eta} - \frac{k+1}{n \eta}}^2} / \p{2 + \frac{\gamma \underset{\bar{}}{\pi}}{\eta} - \frac{k+1}{n \eta}}} \;.
\end{aligned}
\end{equation*}
By going further and imposing that $k + 1 \leq \frac{1}{2} n \gamma \underset{\bar{}}{\pi} $, we get 
\begin{equation*}
\begin{aligned}
    \Prob \p{X_{(E(np) + k)} > F_X^{-1}(p) + \gamma} 
    &\leq e^{- \frac{n \eta}{4} {\p{\frac{\gamma \underset{\bar{}}{\pi}}{\eta}}^2} / \p{2 + \frac{\gamma \underset{\bar{}}{\pi}}{2 \eta}}} \;.
\end{aligned}
\end{equation*}
Finally, by noticing that $\eta {\p{\frac{\gamma \underset{\bar{}}{\pi}}{\eta}}^2} / \p{2 + \frac{\gamma \underset{\bar{}}{\pi}}{2 \eta}} = \frac{\gamma^2 \underset{\bar{}}{\pi}^2}{2 (1-p) - \frac{3}{2} \gamma \underset{\bar{}}{\pi}} \geq \frac{\gamma^2 \underset{\bar{}}{\pi}^2}{2 (1-p)}$,
\begin{equation*}
\begin{aligned}
    \Prob \p{X_{(E(np) + k)} > F_X^{-1}(p) + \gamma} 
    &\leq e^{- \frac{\gamma^2 \underset{\bar{}}{\pi}^2}{8 (1-p)} n} \;.
\end{aligned}
\end{equation*}
Now, looking at the other inequality, we define
\begin{equation*}
    \underset{\bar{}}{N} \eqdef \sum_{i=1}^n \Ind_{(- \infty, F_X^{-1}(p) - \gamma)} (X_i) \;.
\end{equation*}
Like previously, 
\begin{equation*}
    \p{X_{(E(np) + k)} < F_X^{-1}(p) - \gamma}
    \subset \p{\underset{\bar{}}{N} \geq  E(np) + k} 
    \subset \p{\underset{\bar{}}{N} \geq  np + k - 1} \;. 
\end{equation*}
With the exact same techniques as previously, imposing the condition $k-1 \geq - \frac{1}{2} n \gamma \underset{\bar{}}{\pi}$ gives 
\begin{equation*}
\begin{aligned}
    \Prob \p{X_{(E(np) + k)} < F_X^{-1}(p) - \gamma} 
    &\leq e^{- \frac{\gamma^2 \underset{\bar{}}{\pi}^2}{8 p} n} \;.
\end{aligned}
\end{equation*}
Thus, under the various conditions specified for $k$, by union bound,
\begin{equation*}
\begin{aligned}
    \Prob \p{|X_{(E(np) + k)} - F_X^{-1}(p)| > \gamma} 
    &\leq e^{- \frac{\gamma^2 \underset{\bar{}}{\pi}^2}{8 p} n} + e^{- \frac{\gamma^2 \underset{\bar{}}{\pi}^2}{8 (1-p)} n}\;.
\end{aligned}
\end{equation*}
Now define $I \eqdef \{ k \in \{ -E(np), \dots, n - E(np)\} | |X_{(E(np) + k)} - F_X^{-1}(p)| \leq \gamma\}$. Notice that since $X_{(1)} \leq X_{(2)} \leq \dots \leq X_{(n)}$, I is an integer interval. Which means that if $a \in I \leq b \in I$, then $[a, b] \cap \Z \subset I$. As a consequence, if $|X_{(E(np) + k)} - F_X^{-1}(p)| \leq \gamma$ for two integers $k_1$ and $k_2$, it is also the case for all the integers between them. By union bound, we get
\begin{equation*}
\begin{aligned}
    \Prob \p{\sup_{k \in J} |X_{(E(np) + k)} - F_X^{-1}(p)| > \gamma} 
    &\leq 2 e^{- \frac{\gamma^2 \underset{\bar{}}{\pi}^2}{8 p} n} + 2 e^{- \frac{\gamma^2 \underset{\bar{}}{\pi}^2}{8 (1-p)} n}\;,
\end{aligned}
\end{equation*}
where 
\begin{equation*}
    J \eqdef \left\{ \max \p{-E(np) + 1,  - E\p{\frac{1}{2} n \gamma \underset{\bar{}}{\pi}} + 1}, \dots, \min \p{n-E(np),   E\p{\frac{1}{2} n \gamma \underset{\bar{}}{\pi}} - 1} \right\} \;.
\end{equation*}

\section{Proof of \Cref{lemma:gapsconcentration}}

\label{proofoflemma:gapsconcentration}
The following fact is a direct consequence of Lemma 2.1 in Chapter 5 of \cite{Devr86}.
\begin{fact}[Concentration of the gaps for uniform samples]
\label{fact:gapsuniform}
    Let $X_1, \dots, X_n \iid U([0, 1])$, the uniform distribution on $[0, 1]$. Denoting $\Delta_1 \eqdef X_{(1)}, \Delta_2 \eqdef X_{(2)} - X_{(1)}, \dots, \Delta_n \eqdef X_{(n)} - X_{(n-1)}$, and $\Delta_{n+1} \eqdef 1 - X_{(n)}$, for any $\gamma >0$ such that $\gamma < \frac{1}{n+1}$,
    \begin{equation*}
        \Prob \p{\min_i \Delta_i > \gamma} = \p{1 - (n+1) \gamma}^n \;.
    \end{equation*}
\end{fact}
We give a proof here for completeness. The first step consists in characterizing the distribution of $(\Delta_1, \dots, \Delta_n)$. Let $h : \R^n \rightarrow \R$ be a positive Borelian function. By the transfer theorem,
\begin{equation*}
    \int h(\Delta_1, \dots, \Delta_n) d\Prob(\Delta_1, \dots, \Delta_n) = \int h(X_{(1)}, X_{(2)} - X_{(1)}, \dots, X_{(n)} - X_{(n-1)}) d\Prob(X_{(1)}, \dots, X_{(n)}) \;.
\end{equation*}
Furthermore, $(X_{(1)}, \dots, X_{(n)})$ follows a uniform distribution on the set of $n$ ordered points in $[0, 1]$. Hence,
\begin{equation*}
    \int h(\Delta_1, \dots, \Delta_n) d\Prob(\Delta_1, \dots, \Delta_n) = n! \int h(X_{1}, X_{2} - X_{1}, \dots, X_{n} - X_{n-1})\Ind_{0 \leq X_1 \leq \dots \leq X_n \leq 1}  dX_1 \dots dX_n \;.
\end{equation*}
Finally, the variable swap $\delta_1 = X_1, \delta_2 = X_2 - X_1, \dots, \delta_{n} = X_n - X_{n_1}$ that has a jacobian of $1$, same as its inverse (both transformations are triangular matrices with only $1$'s on the diagonal), gives that 
\begin{equation*}
    \int h(\Delta_1, \dots, \Delta_n) d\Prob(\Delta_1, \dots, \Delta_n) = n! \int h(\delta_1,  \dots, \delta_n)\Ind_{0 \leq \delta_1, \dots, 0 \leq \delta_n, \sum_{i=1}^n \delta_i \leq 1}  d\delta_1 \dots d\delta_n \;.
\end{equation*}
The last equation means that $(\Delta_1, \dots, \Delta_n)$ follows a uniform distribution on the simplex $\bigg\{ 0 \leq \Delta_1,  \dots, \Delta_n \leq 1, \sum_{i=1}^n \Delta_i \leq 1\bigg\}$.
The probability $ \Prob \p{\min_i \Delta_i > \gamma}$ may now be computed as
\begin{equation*}
    \begin{aligned}
         \Prob \p{\min_i \Delta_i > \gamma} 
         &=
         n! \int \Ind_{\gamma < \delta_1, \dots, \gamma < \delta_n, \sum_{i=1}^n \delta_i < 1 - \gamma} \Ind_{0 \leq \delta_1, \dots, 0 \leq \delta_n, \sum_{i=1}^n \delta_i \leq 1}  d\delta_1 \dots d\delta_n , 
    \end{aligned}
\end{equation*}
and by considering the variable swap $\delta_i ' \eqdef \frac{\delta_i - \gamma}{1 - (n+1) \gamma}$ (which is separable) of which the jacobian of the inverse is $(1-(n+1)\gamma)^n$,
\begin{equation*}
    \begin{aligned}
         \Prob \p{\min_i \Delta_i > \gamma} 
         &=
         n! (1-(n+1)\gamma)^n \int \Ind_{0 < \delta_1', \dots, 0 < \delta_n', \sum_{i=1}^n \delta_i' < 1}  d\delta_1' \dots d\delta_n' 
         =
         (1-(n+1)\gamma)^n\;.
    \end{aligned}
\end{equation*}
This concludes the proof of \Cref{fact:gapsuniform}.
Now, $X_1, \dots, X_n \iid \Prob_\pi$ where $\pi$ is a density on $[0, 1]$ w.r.t. Lebesgue's measure such that $\bar{\pi} \in \R  \geq \pi \geq \underset{\bar{}}{\pi} \in \R >0$ almost surely. In particular, the data is not necessary uniform. By a coupling argument, if $U_1, \dots, U_n \iid U([0, 1])$, $\p{F_\pi^{-1}(U_1), \dots, F_\pi^{-1}(U_n)}$ has the same distribution as $\p{X_1, \dots, X_n}$. We can furthermore notice that 
\begin{equation*}
    \forall p, q \in (0, 1), \epsilon > 0, \quad, |p - q|>\epsilon \implies \left|F_\pi^{-1}(p) - F_\pi^{-1}(q)\right| > \frac{\epsilon}{\bar{\pi}} \;.
\end{equation*}
Indeed, the lower bound $\pi \geq \underset{\bar{}}{\pi}$ ensures that $F_\pi$ is a bijection and that so does its inverse. The upper bound $\bar{\pi}\geq \pi$ ensures that $F_\pi$ cannot grow too fast, and thus that its inverse is not too flat.
Formally,
\begin{equation*}
    \forall a, b, \quad |F_\pi(b) - F_\pi(a)| = \left| \int_a^b \pi \right| \leq \bar{\pi} |b-a|.
\end{equation*}
In particular, it holds for $b = F_\pi^{-1}(p)$ and $a = F_\pi^{-1}(q)$.

Consequently, if $\Delta_1' \eqdef U_{(1)}, \Delta_2' \eqdef U_{(2)} - U_{(1)}, \dots, \Delta_n' \eqdef U_{(n)} - U_{(n-1)}$, and $\Delta_{n+1}' \eqdef 1 - U_{(n)}$,
\begin{equation*}
    \Prob \p{\min_i \Delta_i > \gamma} 
    \geq 
    \Prob \p{\min_i \Delta_i' >\bar{\pi}  \gamma} 
    = 
    \p{1 - (n+1) \bar{\pi} \gamma}^n \;.
\end{equation*}
Finally, let us simplify this expression to a easy-to-handle one. If $\gamma < \frac{n}{2 \bar{\pi}}$,
\begin{equation*}
    \Prob \p{\min_i \Delta_i > \frac{\gamma}{n^2}} 
    = 
    \p{1 - \frac{n+1}{n} \frac{\bar{\pi} \gamma}{n}}^n 
    \geq
    \p{1 - \frac{2n}{n} \frac{\bar{\pi} \gamma}{n}}^n
    =
    \p{1 - \frac{2 \bar{\pi} \gamma}{n}}^n
    \;.
\end{equation*}
Furthermore, for any $x \in (0, 1/2)$ and $n \geq 1$, by the Taylor-Lagrange formula, there exist $c \in \p{-\frac{x}{n}, 0}$
\begin{equation*}
    \begin{aligned}
        \p{1-\frac{x}{n}}^n 
        &= 
        e^{n \ln \p{1-\frac{x}{n}}} 
        = 
        e^{n \p{-\frac{x}{n} - \frac{1}{2} \frac{1}{(1 + c)^2} \frac{x^2}{n^2} } } 
    \end{aligned}
\end{equation*}
And so, when $n \geq 1$,
\begin{equation*}
    \begin{aligned}
        \p{1-\frac{x}{n}}^n 
        \geq
        e^{-2x}
    \end{aligned}
\end{equation*}
In definitive, when $n \geq 1$ and $\gamma < \frac{1}{4 \bar{\pi}}$
\begin{equation*}
    \Prob \p{\min_i \Delta_i > \frac{\gamma}{n^2}} \geq e^{- 4 \bar{\pi} \gamma} \;.
\end{equation*}

\section{Proof of \Cref{theorem:convergenceqexp}}
\label{proofoftheorem:convergenceqexp}

For simplicity, let us assume that $E\p{\frac{1}{2} n \gamma \underset{\bar{}}{\pi}} - 1 \leq \min \p{E(np) - 1, n - E(np)}$, which is for instance the case when $ \gamma < \frac{2 \min(p, 1-p)}{\underset{\bar{}}{\pi}}$, which we suppose. Furthermore, suppose that $\frac{1}{2} n \gamma \underset{\bar{}}{\pi} \geq 2$, which is for instance the case when $n > 2 / \min(p, 1-p)$ thank to the hypothesis on $\gamma$. 
By noting $K \eqdef E\p{\frac{1}{4} n \gamma \underset{\bar{}}{\pi}}$, \Cref{lemma:concentrationempiricalquantiles} says that,
\begin{equation*}
\begin{aligned}
    \Prob \p{\sup_{k \in \{ -K, \dots, K\}} |X_{(E(np) + k)} - F_X^{-1}(p)| > \gamma} 
    &\leq 4 e^{- \frac{\gamma^2 \underset{\bar{}}{\pi}^2}{8 \max(p,(1-p))} n}\;,
\end{aligned}
\end{equation*}
We call $QC$ (for \emph{quantile concentration}) the complementary of this last event.
Let $\delta >0$ that satisfies $\delta < \frac{1}{4 \bar{\pi}}$. We define the event $G \eqdef \p{\min_i \Delta_i > \frac{\delta}{n^2}}$ (for \emph{gaps}). \Cref{lemma:gapsconcentration} ensures that
\begin{equation*}
    \Prob \p{G^c} \leq 1 - e^{- 4 \bar{\pi} \delta} \;.
\end{equation*}
Conditionally to $QC$, denoting by $q$ the output of QExp, $| q - F_\pi^{-1}(p) | > \gamma \implies \mathfrak{E} \geq K-1 \geq K/2$ whenever $n  \geq 4 / (\gamma \underset{\bar{}}{\pi})$. By also working conditionally to $G$, and in order to apply \Cref{fact:empiricalutilityone}, we look for a $\beta > 0$ such that
\begin{equation*}
    K/2 = 2 \frac{\ln (n^2) + \ln \p{\frac{1}{\delta}} + \ln \p{\frac{1}{\beta}} }{\epsilon} \;,
\end{equation*}
which gives 
\begin{equation*}
    \beta = \frac{n^2}{\delta} e^{- \frac{\epsilon E\p{\frac{1}{4} n \gamma \underset{\bar{}}{\pi} }}{4}} \;.
\end{equation*}
Note that even if \Cref{fact:empiricalutilityone} is stated for $\beta \in (0, 1)$, its conclusion remains obviously true for $\beta \geq 1$.

Finally,
\begin{equation*}
    \begin{aligned}
        \Prob \p{| q - F_\pi^{-1}(p) | > \gamma}
        &\leq 
        \Prob \p{| q - F_\pi^{-1}(p) | > \gamma, QC, G} + \Prob \p{QC^c} +  \Prob \p{G^c} \\
        &\leq   \frac{e n^2}{\delta} e^{- \frac{\epsilon n \gamma \underset{\bar{}}{\pi} }{16}} + 1 - e^{- 4 \bar{\pi} \delta} +  4e^{- \frac{\gamma^2 \underset{\bar{}}{\pi}^2}{8 \max (p, 1-p)} n},
    \end{aligned}
\end{equation*}
and by fixing $\delta \eqdef \frac{n \sqrt{e}}{2 \sqrt{2 \bar{\pi}}}  e^{- \frac{\epsilon n \gamma \underset{\bar{}}{\pi} }{32}}$, because $ 1 - e^{- 4 \bar{\pi} \delta} \leq 8  \bar{\pi} \delta$ for any $\delta > 0$,
\begin{equation*}
    \begin{aligned}
        \Prob \bigg(| q - F_\pi^{-1}(p) | > \gamma\bigg)
        &\leq 
        4 n \sqrt{2 e \bar{\pi}} e^{- \frac{\epsilon n \gamma \underset{\bar{}}{\pi} }{32}} +  4e^{- \frac{\gamma^2 \underset{\bar{}}{\pi}^2}{8\max(p,(1-p))} n} \;.
    \end{aligned}
\end{equation*}

\section{Proof of \Cref{theorem:convergenceindexp}}
\label{proofoftheorem:convergenceindexp}
IndExp is the application of $m$ independent QExp procedures but with privacy parameter $\frac{\epsilon}{m}$ in each. A union bound on the events that check if each quantile is off by at least $\gamma$ gives the result by \Cref{theorem:convergenceqexp}.

\section{Proof of \Cref{theorem:convergencerecexp}}
\label{proofoftheorem:convergencerecexp}

For simplicity, let us assume that $E\p{\frac{1}{2} n \gamma \underset{\bar{}}{\pi}} - 1 \leq \min \p{E(np_1) - 1, n - E(np_m)}$, which is for instance the case when $ \gamma < \frac{2 \min_i \min(p_i, 1-p_i)}{\underset{\bar{}}{\pi}}$, which we suppose. Furthermore, suppose that $\frac{1}{2} n \gamma \underset{\bar{}}{\pi} \geq 2$ , which is for instance the case when $n > 2 / \min_i \min(p_i, 1-p_i)$ thank to the hypothesis on $\gamma$.
By noting $K \eqdef E\p{\frac{1}{4} n \gamma \underset{\bar{}}{\pi}}$, \Cref{lemma:concentrationempiricalquantiles} says that for any $i \in \{ 1, \dots, m\}$,
\begin{equation*}
\begin{aligned}
    \Prob \p{\sup_{k \in \{ -K, \dots, K\}} |X_{(E(np_i) + k)} - F_X^{-1}(p_i)| > \gamma} 
    &\leq 4 e^{- \frac{\gamma^2 \underset{\bar{}}{\pi}^2}{8 C_{p_1, \dots, p_m}} n}\;,
\end{aligned}
\end{equation*}
where $C_{p_1, \dots, p_m} \eqdef \max_i\p{\max\p{p_i, (1-p_i)}}$. We define the event $QC$ (for \emph{quantile concentration}),
\begin{equation*}
    QC \eqdef \bigcap_{i=1}^m \p{\sup_{k \in \{ -K, \dots, K\}} |X_{(E(np_i) + k)} - F_X^{-1}(p_i)| \leq \gamma} \;.
\end{equation*}
By union bounds,
\begin{equation*}
    \Prob \p{QC^c} \leq 4 m e^{- \frac{\gamma^2 \underset{\bar{}}{\pi}^2}{8 C_{p_1, \dots, p_m}} n}\;.
\end{equation*}
Let $\delta >0$ that satisfies $\delta < \frac{1}{4 \bar{\pi}}$. We define the event $G \eqdef \p{\min_i \Delta_i > \frac{\delta}{n^2}}$ (for \emph{gaps}). \Cref{lemma:gapsconcentration} ensures that
\begin{equation*}
    \Prob \p{G^c} \leq 1 - e^{- 4 \bar{\pi} \delta} \;.
\end{equation*}
Conditionally to $QC$, denoting by $\vect{q}$ the output of RecExp, $\| \vect{q} - F_\pi^{-1}(\vect{p}) \|_\infty > \gamma \implies \mathfrak{E} \geq K-1 \geq K/2$ whenever $n  \geq 4 / (\gamma \underset{\bar{}}{\pi})$. By also working conditionally to $G$, and in order to apply \Cref{fact:empiricalutility}, we look for a $\beta > 0$ such that
\begin{equation*}
    K/2 = 2 (\log_2 m + 1)^2 \frac{\ln (n^2) + \ln \p{\frac{1}{\delta}} + \ln m + \ln \p{\frac{1}{\beta}} }{\epsilon} \;,
\end{equation*}
which gives 
\begin{equation*}
    \beta = \frac{n^2 m}{\delta} e^{- \frac{\epsilon E\p{\frac{1}{4} n \gamma \underset{\bar{}}{\pi} }}{4(\log_2 m + 1)^2}} \;.
\end{equation*}
Note that even if \Cref{fact:empiricalutility} is stated for $\beta \in (0, 1)$, its conclusion remains obviously true for $\beta \geq 1$.

Finally,
\begin{equation*}
    \begin{aligned}
        \Prob \p{\| \vect{q} - F_\pi^{-1}(\vect{p}) \|_\infty > \gamma}
        &\leq 
        \Prob \p{\| \vect{q} - F_\pi^{-1}(\vect{p}) \|_\infty > \gamma, QC, G} + \Prob \p{QC^c} +  \Prob \p{G^c} \\
        &\leq  \frac{e n^2 m}{\delta} e^{- \frac{\epsilon  n \gamma \underset{\bar{}}{\pi} }{32 (\log_2 m + 1)^2}} + 1 - e^{- 4 \bar{\pi} \delta} +  4 m e^{- \frac{\gamma^2 \underset{\bar{}}{\pi}^2}{8 C_{p_1, \dots, p_m}} n},
    \end{aligned}
\end{equation*}
and by fixing $\delta \eqdef \frac{n \sqrt{ e m}}{2 \sqrt{2 \bar{\pi}}}  e^{- \frac{\epsilon  n \gamma \underset{\bar{}}{\pi} }{32 (\log_2 m + 1)^2}}$, we get that,
\begin{equation*}
    \begin{aligned}
        \Prob \p{\| \vect{q} - F_\pi^{-1}(\vect{p}) \|_\infty > \gamma}
        &\leq 
        4 n \sqrt{2 e \bar{\pi} m} e^{- \frac{\epsilon  n \gamma \underset{\bar{}}{\pi} }{32 (\log_2 m + 1)^2}} +  4 m e^{- \frac{\gamma^2 \underset{\bar{}}{\pi}^2}{8 C_{p_1, \dots, p_m}} n} \;.
    \end{aligned}
\end{equation*}

\section{Proof of \Cref{lemma:lowerboundqexp}}
\label{proofoflemma:lowerboundqexp}
By definition of $u_{\text{QExp}}$ we have $-n \leq
u_{\text{QExp}} \big( (X_1, \dots, X_n), q \big) \leq 0$ for any input, hence using that $0 \leq \gamma \leq 1/4$ we get
\begin{equation*}
    \begin{aligned}
        \Prob \bigg(| q - t | > \gamma\bigg) 
        &= \frac{\int_{
        [0, 1] \backslash [t-\gamma,t+\gamma]
        } e^{\frac{\epsilon}{2} u_{\text{QExp}} \big( (X_1, \dots, X_n), q \big) } dq}{\int_{[0, 1]} e^{\frac{\epsilon}{2} u_{\text{QExp}} \big( (X_1, \dots, X_n), q \big) } dq} \\
        &\geq \frac{\int_{
         [0, 1] \backslash [t-\gamma,t+\gamma]
} e^{-\frac{\epsilon}{2}n} dq}{\int_{[0, 1]} e^{0} dq} \\
        &\geq (1-2 \gamma) e^{-\frac{\epsilon}{2}n} \\
        &\geq \frac{1}{2} e^{-\frac{\epsilon}{2}n} \;.
    \end{aligned}
\end{equation*}

\section{Proof of \Cref{lemma:histogramconvergencedensity}}
\label{proofoflemma:histogramconvergencedensity}

Let us consider a specific bin of the histogram $b$. Let $\gamma > 0$. Denoting by $\|\cdot \|_{\infty, b}$ the infinite norm restrained to the support of $b$, which is a semi-norm, we have 
\begin{equation*}
    \begin{aligned}
        \Prob \p{\|\hat{\pi}^{\text{hist}} - \pi \|_{\infty, b} > \gamma}
        &=
        \Prob \p{\left\| \frac{1}{nh} \p{\sum_{i=1}^n \Ind_{b}(X_i) + \frac{2}{\epsilon} \mathcal{L}}- \pi \right\|_{\infty, b} > \gamma}
    \end{aligned}
\end{equation*}
where $\mathcal{L} \sim \text{Lap}(1)$, a centered Laplace distribution of parameter $1$. So,
\begin{equation*}
    \begin{aligned}
        \Prob \p{\|\hat{\pi}^{\text{hist}} - \pi \|_{\infty, b} > \gamma}
        &=
        \Prob \p{\left\| \p{\frac{1}{nh} \sum_{i=1}^n \Ind_{b}(X_i) - \pi} + \frac{2}{n h \epsilon} \mathcal{L} \right\|_{\infty, b} > \gamma} \\
        &\stackrel{\text{triangular inequality}}{\leq}
        \Prob \p{\left\| \frac{1}{nh} \sum_{i=1}^n \Ind_{b}(X_i) - \pi \right\|_{\infty, b} > \gamma/2} + \Prob \p{\left|\frac{2}{n h \epsilon} \mathcal{L} \right| > \gamma/2}
    \end{aligned}
\end{equation*}
Let us first control the first term. Since $\pi$ is $L$ Lipschitz, $\forall x \in b, \left| \pi(x) - \frac{1}{h} \int_b \pi\right| \leq \frac{Lh}{2}$. So, when $\gamma > Lh$,
\begin{equation*}
    \begin{aligned}
        \p{\left\| \frac{1}{nh} \sum_{i=1}^n \Ind_{b}(X_i) - \pi \right\|_{\infty, b} > \gamma/2}
        \subset 
        \p{\left| \frac{1}{nh} \sum_{i=1}^n \Ind_{b}(X_i) - \frac{1}{h} \int_b\pi \right| > \gamma/2 - Lh/2} \;.
    \end{aligned}
\end{equation*}
Finally, notice that the family $\p{\Ind_{b}(X_i)}_i$ is a family of i.i.d. Bernoulli random variables of probability of success $\int_b \pi$. By Hoeffding's inequality,
\begin{equation*}
    \Prob \p{\left\| \frac{1}{nh} \sum_{i=1}^n \Ind_{b}(X_i) - \pi \right\|_{\infty, b} > \gamma/2} 
    \leq 
    2 e^{- \frac{h^2 (\gamma - Lh)^2}{4} n} \;.
\end{equation*}
The second term is controlled via a tail bound on the Laplace distribution as
\begin{equation*}
    \begin{aligned}
        \Prob \p{\left|\frac{2}{n h \epsilon} \mathcal{L} \right| > \gamma/2} 
        &= \Prob \p{\left|\mathcal{L} \right| > \frac{\gamma n h \epsilon}{4}} \\
        &= \int_{\frac{\gamma n h \epsilon}{4}}^{\infty} e^{-t} dt \\
        &= e^{- \frac{\gamma h n \epsilon}{4}} \;.
    \end{aligned}
\end{equation*}
So, if $\gamma > Lh$, 
\begin{equation*}
    \Prob \p{\|\hat{\pi}^{\text{hist}} - \pi \|_{\infty, b} > \gamma} \leq 2 e^{- \frac{h^2 (\gamma - Lh)^2}{4} n} + e^{- \frac{\gamma h n \epsilon}{4}} \;.
\end{equation*}
Finally, a union bound on all the bins gives that if $\gamma > Lh$,
\begin{equation*}
    \Prob \p{\|\hat{\pi}^{\text{hist}} - \pi \|_{\infty} > \gamma} \leq \frac{2}{h} e^{- \frac{h^2 (\gamma - Lh)^2}{4} n} + \frac{1}{h} e^{- \frac{\gamma h n \epsilon}{4}} \;.
\end{equation*}

\section{Proof of \Cref{lemma:densityesttoqfest}}
\label{proofoflemma:densityesttoqfest}

We have,
    \begin{equation*}
        \begin{aligned}
            F_{\hat{\pi}} \p{F^{-1}_{\pi} (p) + \frac{\alpha}{\underset{\bar{}}{\pi}}}
            &\stackrel{\left\| \hat{\pi} - \pi \right\|_{\infty}  \leq \alpha}{\geq }
            F_{\pi} \p{F^{-1}_{\pi} (p) + \frac{\alpha}{\underset{\bar{}}{\pi}}} - \alpha \\
            &\stackrel{\pi \geq \underset{\bar{}}{\pi}}{\geq }
            F_{\pi} \p{F^{-1}_{\pi} (p)} + \frac{\alpha}{\underset{\bar{}}{\pi}} \underset{\bar{}}{\pi} - \alpha \\
            &= F_{\pi} \p{F^{-1}_{\pi} (p)} = p \;.
        \end{aligned}
    \end{equation*}
    So, 
    \begin{equation*}
        F^{-1}_{\hat{\pi}} (p) \leq F^{-1}_{\pi} (p) + \frac{\alpha}{\underset{\bar{}}{\pi}} \;.
    \end{equation*}
    Furthermore, for any $t \in \left[\frac{2\alpha}{\underset{\bar{}}{\pi}}, F^{-1}_{\pi} (p) \right]$,
    \begin{equation*}
        \begin{aligned}
            F_{\hat{\pi}} \p{F^{-1}_{\pi} (p) - t}
            &\stackrel{\left\| \hat{\pi} - \pi \right\|_{\infty}  \leq \alpha}{\leq }
            F_{\pi} \p{F^{-1}_{\pi} (p) - t} + \alpha \\
            &\stackrel{\pi \geq \underset{\bar{}}{\pi}}{\leq }
            F_{\pi} \p{F^{-1}_{\pi} (p)} - t \underset{\bar{}}{\pi} + \alpha \\
            &<
            F_{\pi} \p{F^{-1}_{\pi} (p)} - \frac{2\alpha}{\underset{\bar{}}{\pi}} \underset{\bar{}}{\pi} + \alpha \\
            &= F_{\pi} \p{F^{-1}_{\pi} (p)} - \alpha < p \;.
        \end{aligned}
    \end{equation*}
    So, for any $t \in \p{\frac{2\alpha}{\underset{\bar{}}{\pi}}, F^{-1}_{\pi} (p)}$;
    \begin{equation*}
        F^{-1}_{\hat{\pi}} (p) \geq F^{-1}_{\pi} (p) - t \;,
    \end{equation*}
    and finally, 
    \begin{equation*}
        F^{-1}_{\hat{\pi}} (p) \geq F^{-1}_{\pi} (p) - \frac{2\alpha}{\underset{\bar{}}{\pi}} \;.
    \end{equation*}

\section{Proof of \Cref{theorem:convergencehistogram}}
\label{proofoftheorem:convergencehistogram}

Given $\gamma \in \p{\frac{2 L h}{\underset{\bar{}}{\pi}}, \gamma_0}$, $\frac{\gamma \underset{\bar{}}{\pi}}{2} \geq \frac{2 \underset{\bar{}}{\pi} L h}{2 \underset{\bar{}}{\pi}} = L h$. So, \Cref{lemma:histogramconvergencedensity} applies and gives that 
\begin{equation*}
    \Prob \p{\|\hat{\pi}^{\text{hist}} - \pi \|_{\infty} > \frac{\gamma \underset{\bar{}}{\pi}}{2}} \leq 
    \frac{1}{h} e^{- \frac{\gamma \underset{\bar{}}{\pi} h n \epsilon}{8}} + \frac{2}{h} e^{- \frac{h^2 }{4} \p{\frac{\gamma \underset{\bar{}}{\pi}}{2} - Lh}^2 n} \;.
\end{equation*}
Furthermore, $I = F_\pi \p{(\gamma_0, 1-\gamma_0}$. So,
\begin{equation*}
    \forall p \in I, \quad \gamma_0 < F_\pi^{-1}(p) < 1 - \gamma_0 \;.
\end{equation*}
In particular, when $\hat{\pi}^{\text{hist}}$ satisfies $\| \hat{\pi}^{\text{hist}} - \pi\| \leq \frac{\gamma \underset{\bar{}}{\pi}}{2}$, \Cref{lemma:densityesttoqfest} applies and gives 
\begin{equation*}
    \forall p \in I, \quad |F_{\hat{\pi}^{\text{hist}}}^{-1}(p) - F_{\pi}^{-1}(p)| \leq \gamma \;.
\end{equation*}
This is equivalent to 
\begin{equation*}
    \forall p \in I, \quad \|F_{\hat{\pi}^{\text{hist}}}^{-1}(p) - F_{\pi}^{-1}(p)\|_{\infty, I} \leq \gamma \;.
\end{equation*}
Finally, 
\begin{equation*}
        \begin{aligned}
            \Prob \bigg( & \|F_{\hat{\pi}^{\text{hist}}}^{-1} - F_{\pi}^{-1}\|_{\infty, I} > \gamma \bigg) \leq  \frac{1}{h} e^{- \frac{\gamma \underset{\bar{}}{\pi} h n \epsilon}{8}} + \frac{2}{h} e^{- \frac{h^2 }{4} \p{\frac{\gamma \underset{\bar{}}{\pi}}{2} - Lh}^2 n} \;.
        \end{aligned} 
    \end{equation*}
\section{Distributions for the experiments}
\label{sec:distributions}

\begin{figure}[h!]
    \centering
    \includegraphics[width=0.6\linewidth]{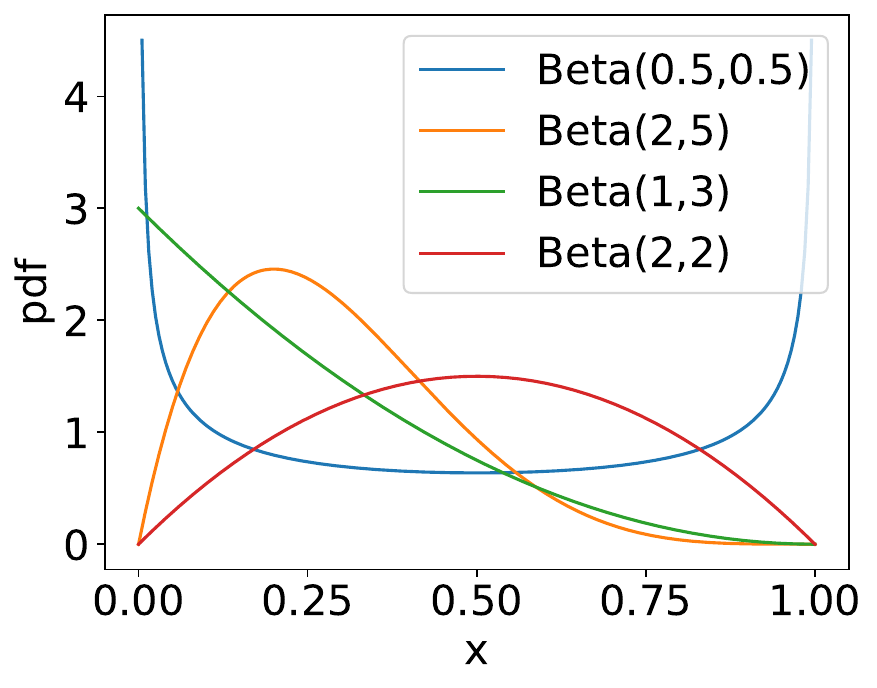} \\
    \centering 
    pdf : "probability distribution function", is the density w.r.t. Lebesgue's measure.
    \caption{Distributions used for the experiments}
    \label{fig:distributions}
\end{figure}


\end{document}